\documentclass[sigconf]{acmart}
\usepackage[ruled,vlined,linesnumbered]{algorithm2e}
\usepackage{multirow, multicol}
\newcommand{\oursys}{\texttt{HARMONY}\xspace}
\newtheorem{myDef}{Definition}
\usepackage{float}
\usepackage{graphicx}
\usepackage{subfigure}

\begin{document}




\title{Adaptive Two-Stage Cloud Resource Scaling via Hierarchical Multi-Indicator Forecasting and Bayesian Decision-Making}


\settopmatter{authorsperrow=4}
\author{Yang Luo}
\email{floating-dream@sjtu.edu.cn}
\affiliation{
  \institution{Shanghai Jiao Tong Uni.}
  \city{Shanghai}
  \country{China}
}

\author{Shiyu Wang}
\email{weiming.wsy@antgroup.com}
\affiliation{%
  \institution{AntGroup}
  \city{Hangzhou}
  \country{China}
}

\author{Zhemeng Yu}
\email{fish_meng@sjtu.edu.cn}
\affiliation{
  \institution{Shanghai Jiao Tong Uni.}
  \city{Shanghai}
  \country{China}
}

\author{Wei Lu}
\email{xiaobo.lw@antgroup.com}
\affiliation{%
  \institution{AntGroup}
  \city{Hangzhou}
  \country{China}
}

\author{Xiaofeng Gao}

\authornote{Xiaofeng Gao is the corresponding author.}
\email{gao-xf@cs.sjtu.edu.cn}
\affiliation{
  \institution{Shanghai Jiao Tong Uni.}
  \city{Shanghai}
  \country{China}
}

\author{Lintao Ma}
\email{maining.mlt@antgroup.com}
\affiliation{%
  \institution{AntGroup}
  \city{Hangzhou}
  \country{China}
}

\author{Guihai Chen}
\email{gchen@cs.sjtu.edu.cn}
\affiliation{
  \institution{Shanghai Jiao Tong Uni.}
  \city{Shanghai}
  \country{China}
}


\renewcommand{\shortauthors}{Yang Luo et al.}

\begin{abstract}
The surging demand for cloud computing resources, driven by the rapid growth of sophisticated large-scale models and data centers, underscores the critical importance of efficient and adaptive resource allocation. As major tech enterprises deploy massive infrastructures with thousands of GPUs, existing cloud platforms still struggle with low resource utilization due to key challenges: capturing hierarchical indicator structures, modeling non-Gaussian distributions, and decision-making under uncertainty. To address these challenges, we propose \oursys, an adaptive \underline{H}ierarchical \underline{A}ttention-based \underline{R}esource \underline{M}odeling and Decisi\underline{on}-Making S\underline{y}stem. \oursys combines hierarchical multi-indicator distribution forecasting and uncertainty-aware Bayesian decision-making. It introduces a novel hierarchical attention mechanism that comprehensively models complex inter-indicator dependencies, enabling accurate predictions that can adapt to evolving environment states. By transforming Gaussian projections into adaptive non-Gaussian distributions via Normalizing Flows. Crucially, \oursys leverages the full predictive distributions in an adaptive Bayesian process, proactively incorporating uncertainties to optimize resource allocation while robustly meeting SLA constraints under varying conditions. Extensive evaluations across four large-scale cloud datasets demonstrate \oursys's state-of-the-art performance, significantly outperforming nine established methods. A month-long real-world deployment validated \oursys's substantial practical impact, realizing over 35,000 GPU hours in savings and translating to \$100K+ in cost reduction, showcasing its remarkable economic value through adaptive, uncertainty-aware scaling. 
Our code is available at \url{https://github.com/Floating-LY/HARMONY1}.
\end{abstract}

\settopmatter{printacmref=false}
\setcopyright{none}
\renewcommand\footnotetextcopyrightpermission[1]{}




\maketitle

\section{Introduction}
The rapid growth and increasing complexity of cloud service data centers have amplified the demand for computational resources, underscoring the critical importance of efficient resource allocation. As major technological enterprises expand their infrastructure—such as Meta deploying an AI cloud supercluster with over 24,000 NVIDIA H100 GPUs \cite{meta} and OpenAI scaling to 7,500 GPU servers \cite{openAI}—the scale and management complexity of these data centers have surged considerably. This expansion has resulted in significantly more intricate architectures and operational requirements, leading to challenges in optimizing computational resource utilization \cite{IJCAI-1,Ast}. Consequently, many methods have been proposed to address these scaling issues by analyzing future workloads and making informed resource allocation decisions, aiming to improve resource utilization and economic benefits \cite{Scaling, firm, IPOC, VLDB}. Tackling the resource scaling challenge has become increasingly crucial as cloud service data centers continue to expand in scale.
\begin{figure}
    \centering
    \includegraphics[width=0.9\linewidth]{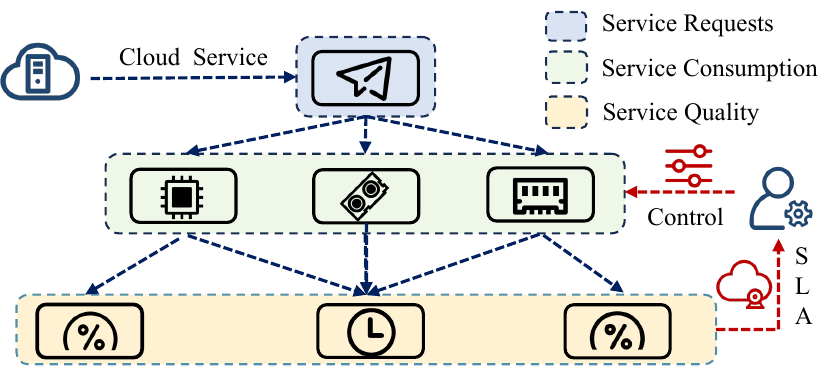}
    \caption{Hierarchical Architecture for Cloud Services. The figure illustrates the hierarchical relationship from service requests, resource consumption (CPU, GPU, Memory) to service quality (Resource Utilization, Response Time). Service provider monitor the service quality to promptly adjust resource consumption for cloud services.}
    \label{fig:intro}
\end{figure}

Although cloud resource scaling remains an actively researched topic, the growing scale and complexity of cloud service continue to present new challenges.
Based on our real-world observations and an extensive literature review, we have identified the following three key challenges for cloud service resource scaling:

\textbf{Challenge 1: Hierarchical Structure in Indicators.} Considering multiple performance indicators simultaneously is imperative for cloud service resource scaling decisions \cite{Magic,group}. For instance, when scaling GPU services, service provider must account for the joint impact of CPU, GPU consumption, and service quality indicators like response time. Single-indicator predictions are insufficient due to the intricate hierarchical relationships among these indicators. As shown in Fig. \ref{fig:intro} and study \cite{causal}, this hierarchy stems from the user access flow: \textbf{service requests} generate computational tasks \textbf{consuming} system resources (e.g., CPU, GPU), which affect the perceived \textbf{service quality}. Relying on any single indicator fails to capture the complex interplay within this hierarchical structure. Accurately understanding and modeling these intricate relationships is crucial for capturing the intrinsic system behavior. 

\textbf{Challenge 2: Non-Gaussian Distribution of Indicator.} 
Cloud services experience frequent business adjustments and complex workloads, leading to non-Gaussian distributions of performance indicators \cite{WSDM, NonG}. These indicators often exhibit non-parametric distributions, reflecting the increasing complexity of their temporal patterns. Traditional Gaussian models are inadequate for accurately capturing the true distributions of cloud service indicators. Thus, more flexible and sophisticated modeling approaches are necessary to avoid inefficiencies and potential violations of Service-Level Agreements (SLAs) caused by inaccurate distribution assumptions.

\textbf{Challenge 3: Decision-making under Uncertainty} 
As cloud service architectures grow more complex, decision-making must consider inherent uncertainties. \cite{Magic, uncertainty} Relying on point estimates can mislead the decision process, especially when prediction accuracies vary across indicators. For example, if GPU usage predictions are less accurate, scaling decisions may become skewed, leading to inefficiencies or SLA violations. The dynamic nature of cloud environments further complicates point prediction, as accuracy can vary significantly across different scenarios. This variability limits the adaptability of point predictions and increases the risk of poor decisions. To mitigate these issues, it's crucial to model uncertainties and integrate them into the decision-making process. 


However, existing methods overlook these key challenges posed by large-scale cloud services. Most works focus on single indicator patterns \cite{IJCAI-1,IPOC,KDD22,kaeinformer,Robust}, neglecting hierarchical inter-indicator relationships (\textbf{Challenge 1}), which limits their adaptivity to evolving business needs. Recent studies \cite{group,Ast} exploring these relationships still rely on point predictions, failing to model decision uncertainties (\textbf{Challenge 3}). While MagicScaler \cite{Magic} combines distribution prediction and optimization for uncertainty modeling, it assumes Gaussian distributions and inadequately captures hierarchical structures, compromising accuracy (\textbf{Challenge 1, 2}). Consequently, uncertainty-aware decision methods leveraging multi-indicator distribution forecasts, which hold significant potential, remain largely underexplored. Appx. \ref{sec: RW} provides further details.

Therefore, we propose the \underline{H}ierarchical \underline{A}ttention-based \underline{R}esource \underline{M}odeling and Decisi\underline{on}-Making S\underline{y}stem (\oursys), a novel unified framework that holistically addresses these challenges. \oursys comprises two main components: Hierarchical Multi-Indicator Distribution Forecasting and Bayesian Decision-Making. The Forecasting component adopts a hierarchical attention mechanism to capture intricate, causal relationships among multiple cloud service indicators (\textbf{Challenge 1}), enabling comprehensive awareness of the evolving environment state. Furthermore, \oursys models the non-Gaussian nature of indicators using Normalizing Flows (\textbf{Challenge 2}), ensuring predictions reliably capture real-world complexities. For decision-making, \oursys integrates the predicted multi-indicator distributions to manage uncertainties dynamically (\textbf{Challenge 3}). This adaptive Bayesian algorithm leverages the full predictive distributions, incorporating uncertainties to optimize resource allocation under varying conditions while meeting SLA constraints. By combining multi-indicator awareness and uncertainty modeling, \oursys achieves highly adaptive scaling that can proactively respond to environment changes and mitigate potential risks. In summary, \oursys fills a critical research gap in large-scale cloud services by proposing a unified framework that:
\begin{itemize}
   \item Introduces a novel hierarchical attention mechanism to capture complex inter-indicator dependencies, addressing causality and improving prediction accuracy.

 \item Models non-Gaussian indicator distributions by Normalizing Flows, capturing real-world cloud indicator complexities for more reliable resource predictions.

 \item Designs the Bayesian decision algorithm that leverages predictive distributions, incorporating uncertainties for optimized resource allocation under SLA constraints.

\item  Achieves superior performance on four large-scale cloud datasets, significantly outperforming nine established methods. In a month-long scaling test, \oursys demonstrated substantial economic impact, saving over 35,000 GPU hours.
\end{itemize}

\section{Problem Definition }
\label{sec: Def}
In this section, we introduce the fundamental concepts of Cloud Service Resource Scaling (CRS). Our task is centered around two objectives: \textbf{Forecasting} the operating conditions of cloud services by analyzing performance indicators.
\textbf{Decision} on the appropriate number of resources for cloud services, guided by the outcomes of forecasting task. Table \ref{tab:symbol} offers a comprehensive overview of the key symbols and their meanings utilized throughout our discussion.

Cloud services are characterized by multiple indicators, each recorded at distinct time stamps and capturing different aspects of the service's performance and utilization. We formally define these historical indicators in Def. ~\ref{Def: Indicators}:

\begin{myDef}[\textbf{Historical Indicators}]
Historical indicators represent the status of a service at different time stamps, including information on service \textbf{requests}, \textbf{consumption}, and \textbf{quality}. The historical indicators are denoted as $\boldsymbol{X} \in \mathbb{R}^{N \times T}$, where $N$ and $T$ denote the number of indicators and time stamps, respectively. In addition, We use $\boldsymbol{X}_r \in \mathbb{R}^{R \times T}$, $\boldsymbol{X}_c \in \mathbb{R}^{C \times T}$, and $\boldsymbol{X}_q \in \mathbb{R}^{Q \times T}$ denotes the service requests, usage, quality indicators where $R+C+Q=N$.
\label{Def: Indicators}
\end{myDef}
Due to the hierarchical nature of cloud service indicators, we introduce hierarchical level information to capture and improve predictive performance. The level information is structured as Def. ~\ref{Def: level}:

\begin{myDef}[\textbf{Level Information}]
Level information contains the hierarchical details of the historical indicators. It is denoted by an array $\boldsymbol{L} \in \mathbb{Z}^{N}$, where each element $L_n \in \{1,\dots, L_{max}\}$ represents the level of the corresponding indicator $X_n$ in the hierarchy. 
\label{Def: level}
\end{myDef}

Based on historical indicators and level information, we aim to predict future distributions of these indicators. The formal definition of the forecasting task is given in Def. ~\ref{Def: Forecasting}:

\begin{myDef}[\textbf{Forecasting}]
Given historical indicators $\boldsymbol{X} \in \mathbb{R}^{N \times T}$ and level information $\boldsymbol{L} \in \mathbb{Z}^{N}$, the goal of forecasting is to predict the values of the indicators at a future time stamp $T+1$. This prediction is denoted as $\boldsymbol{D} \in \mathbb{R}^{N \times S}$, where $S$ is the number of possible states in the distribution of each indicator.
\label{Def: Forecasting}
\end{myDef}

Then we define the basic computational unit for scheduling in cloud services, which may be referred to by different names depending on the system, such as PODs in Kubernetes.

\begin{myDef}[\textbf{Computational Unit}]
A computational unit (CU) comprises a combination of computing resources such as CPU, GPU, and Memory. The computational units have specific configuration denoted as $\boldsymbol{O} \in \mathbb{Z}^{C}$, where each element $O_c$ represents the quantity of the $c$-th resource type. Here $C$ corresponds to the number of service consumption indicators. All computational units for a given cloud service share identical configurations.
\label{Def: POD}
\end{myDef}

After getting the future distributions of service indicators and computational unit configuration, we determine the optimal number of CUs required. The formal definition is provided in Def.~\ref{Def: Decision}:

\begin{myDef}[\textbf{Decision}]
Given the predicted distributions $\boldsymbol{D}$ and computational unit  configuration $\boldsymbol{O}$, the goal of the auto-scaling decision algorithm $\mathcal{A}$ is to determine the appropriate number of CUs $N$ to allocate. This optimal number is denoted by $N_{optimal} = \mathcal{A}(\boldsymbol{D}, \boldsymbol{O})$.
\label{Def: Decision}
\end{myDef}

 \begin{figure*}[htbp]
     \centering
    \includegraphics[width=0.95\linewidth]{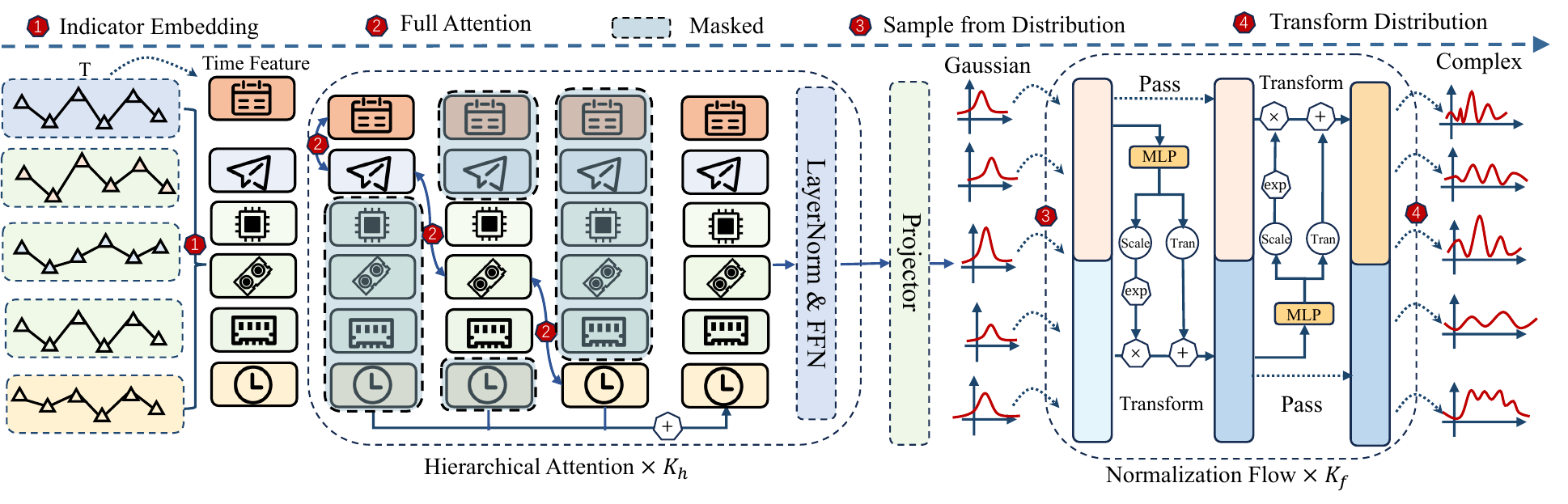}
    \caption{Forecasting Framework. Multiple indicators of cloud services are initially input for indicator embedding, followed by hierarchical attention to capture structural relationships and projection into Gaussian distributions. Finally, the Gaussian distribution is transformed by Normalizing Flows to obtain the distribution that conforms to the indicators.}
    \label{fig:framework}
 \end{figure*}

\section{FORECASTING FUTURE DISTRIBUTIONS}
In this section, we elucidate \oursys for predicting future distributions from historical indicators. Our forecasting model adopts an encoder-decoder architecture. First, the historical indicators are processed through Indicator Embedding within the Encoder to integrate temporal information. Hierarchical Attention is then applied across the embeddings of different indicators to capture their hierarchical relationships. This output is subsequently projected onto a Gaussian distribution. The Decoder part then applies Normalizing Flows to transform the Gaussian distribution into complex, non-parametric distributions. The architecture of the model is depicted in Fig. \ref{fig:framework}. In the following parts of this section, we will introduce  the design of encoder and decoder in detail.

\subsection{Encoder: Capturing Temporal and Hierarchical Dependencies}
The encoder processes historical indicators to extract temporal and hierarchical dependencies, addressing the complex nature of cloud service data highlighted in \textbf{Challenge 1}. Utilizing Indicator Embedding and Hierarchical Attention mechanisms, it captures the diverse, interrelated aspects of cloud services. The encoder then projects these processed embeddings into latent Gaussian distributions, forming a solid basis for predictions. This approach effectively tackles the challenges presented by the data's temporal dynamics and hierarchical structure, ensuring a comprehensive representation of cloud service indicators.

\subsubsection{Indicator Embedding}
The encoder begins with the input of historical indicators $\boldsymbol{X} \in \mathbb{R}^{N \times T}$. Recent research \cite{iTransformer} has demonstrated that embedding along the variable dimension $N$, rather than the temporal dimension $T$, can more effectively model the relationships between variables. This approach is particularly beneficial in large-scale cloud service scenarios where the interdependencies between different indicators are both significant and complex.

To further enhance the temporal dependencies, we extract time features and concatenate them with the indicator embeddings, forming an enhanced feature matrix $\tilde{\boldsymbol{X}} \in \mathbb{R}^{(N+F) \times T}$, where $F$ is the number of time features. For instance, time stamp information such as year, month, and day can be concatenated with the indicators.

Subsequently, the enhanced indicators are projected to the embeddeding $\boldsymbol{E} \in \mathbb{R}^{(N + F) \times D}$ through a dense layer, where $D$ is the dimensionality of the embedding. This projection facilitates the robust extraction of temporal patterns, laying a strong foundation for the Hierarchical Attention Mechanism.

\subsubsection{Hierarchical Attention}
We design a series of blocks for hierarchical attention, each comprising multiple levels.  The input to the $k$-th block is denoted as $\boldsymbol{H}^{k} \in \mathbb{R}^{(N+F) \times D}$.  In each attention block, we implement $L_{max}$ layers of attention to model the structure, where $L_{max}$ is the largest level of indicators.  To incorporate hierarchical information, we introduce level information $\boldsymbol{L} \in \mathbb{R}^{N \times D}$ and expand it into $\tilde{\boldsymbol{L}} \in \mathbb{R}^{(N+F) \times D}$, where we set the layer of time features $\tilde{L}_{N+1}, \dots, \tilde{L}_{N+F}$ as $1$ to capture the global temporal dynamics.  We project the input $\boldsymbol{H}^{k}$ into $\boldsymbol{Q},\boldsymbol{K},\boldsymbol{V} \in \mathbb{R}^{(N+F) \times D}$ for shared use. This approach promotes the sharing and reuse of learned representations and enhance the modeling capability.
 
For the $l$-th layer in the block, it focuses on elements within the same hierarchical level and the output of the preceding layer. Initially, we compute the mask $\boldsymbol{M}_l$ for the current level by Eqn. (\ref{eqn: M}):

\begin{equation}
\boldsymbol{M}_l = \left(\tilde{\boldsymbol{L}} == l\right) \in \mathbb{R}^{(N+F) \times D}
\label{eqn: M}
\end{equation}

Then, we obtain masked queries $\boldsymbol{Q}_l$, keys $\boldsymbol{K}_l$, and values $\boldsymbol{V}_l$ by element-wise multiplication of $\boldsymbol{M}_l$, subsequently adding the output of the previous layer as Eqn. (\ref{eqn: qkv}) denotes. This operation enables the attention mechanism at this layer to focus on the relationships between elements within the current level of indicators, as well as the comprehensive information involving the current level of indicators and those from the preceding layer. Eqn. (\ref{eqn: H}) exhibits the calculating process of layer output $\boldsymbol{H}^{k}_{l+1}$.

\begin{equation}
    \boldsymbol{Q_l}=\boldsymbol{Q}\odot\boldsymbol{M}_l + \boldsymbol{H}^{k}_{l}\,,  \quad 
\boldsymbol{K_l}=\boldsymbol{K}\odot\boldsymbol{M}_l + \boldsymbol{H}^{k}_{l}\,, \quad 
\boldsymbol{V_l}=\boldsymbol{V}\odot\boldsymbol{M}_l + \boldsymbol{H}^{k}_{l}. \label{eqn: qkv}
\end{equation}
\begin{equation}
    \boldsymbol{H}^{k}_{l+1} = Attention(\boldsymbol{Q_l}, \boldsymbol{K_l},\boldsymbol{V_l})\odot\boldsymbol{M}_l. \label{eqn: H}
\end{equation}

The function $Attention(\cdot)$ can take various forms, such as scaled dot-product attention, depending on the specific architecture. In our implementation, we employ the commonly used FullAttention, which calculates the attention score across all the unmasked indicators. By adhering to the hierarchical structure, this mechanism allows each layer to specialize in capturing relationships at a specific level of the data hierarchy. It ensures that no information from lower layers leaks into the higher layers, preserving the  causal order inherent in the data. This design is crucial for accurately modeling inter-indicator dependencies in forecasting task.

Finally, we combine the output of each layer after Layer Normalization \cite{layernorm} and a Feed Forward Network (FFN) as Eqn. (\ref{eqn: FFN}) denotes, which is the commonly-used configuration in transformers. In our implementation, the FFN is implemented by two dense layers. The combined output serves as the input for the next block.
\begin{equation}
    \boldsymbol{H}^{k+1}=FFN(LayerNorm(\sum\nolimits_{l=1}^{L_{max}}\boldsymbol{H}^{k}_{l+1})).  \label{eqn: FFN}
\end{equation}

\subsubsection{Distribution Projector}
Having comprehensively captured the temporal and hierarchical dependencies of the indicators, our objective is to estimate the future conditions of the indicators through a distributional approach. To achieve this, we map the output of the last attention block through a projector to a parameterized Gaussian distribution, which serves as the output of the encoder, as shown in Eqn. (\ref{eqn: E}). $\boldsymbol{H}^{K_h+1}_{:N} \in \mathbb{R}^{N\times D}$ represents the embeddings of the indicators and $Projector: \mathbb{R}^{N\times D} \rightarrow \mathbb{R}^{N\times 2}$ is implemented by two dense layers. The last two dimensions of $\boldsymbol{D}_G$ respectively represent the mean and variance of the Gaussian distribution.
\begin{equation}
    \boldsymbol{D}_G=Projector(\boldsymbol{H}^{K_h}_{:N}) \in \mathbb{R}^{N\times 2}.\label{eqn: E}
\end{equation}

\subsection{Decoder: Transforming Gaussian to Complex Distributions}
As discussed in \textbf{Challenge 2}, the complex distribution of indicators in cloud services renders simple Gaussian models inadequate. Hence, we need to transform the parameterized Gaussian distribution from the encoder into a non-parametric distribution with greater representational flexibility.
To achieve this, we employ a Real-NVP \cite{RealNVP} architecture, a form of Normalizing Flows (NF). It enables efficient sampling and density estimation by transforming simple distributions, such as isotropic Gaussians, into more complex ones through a series of invertible transformations. 

\subsubsection{Normalizing Flow}
The transformation starts by sampling a latent vector $\boldsymbol{Z} \in \mathbb{R}^{N }$ from the Gaussian distribution $\boldsymbol{D}_G$. Each NF consists of two coupling layers and each layer transforms one half of the input embedding while leaving the other half unchanged. Let $\boldsymbol{Z}^k$ denote the input to the $k$-th flow. It is split into two halves: $\boldsymbol{Z}^k_{1} = \boldsymbol{Z}^k_{:\lfloor N/2 \rfloor}$ and $\boldsymbol{Z}^k_{2} = \boldsymbol{Z}^k_{\lfloor N/2 \rfloor:}$.

In the first coupling layer, $\boldsymbol{Z}^k_{1}$ is processed by two separate Multi-Layer Perceptrons (MLPs) to produce the scaling factor and the translation factor  as Eqn. (\ref{eqn: MLP}) describes:
\begin{equation}
\boldsymbol{scale} = MLP_s(\boldsymbol{Z}^k_{1}), \quad \boldsymbol{tran} = MLP_t(\boldsymbol{Z}^k_{1}).\label{eqn: MLP}
\end{equation}
These factors are then used to transform the second half of the input embedding $\boldsymbol{Z}^k_{2}$ in Eqn. (\ref{eqn: Z}). 
\begin{equation}
\tilde{\boldsymbol{Z}}^k_{2} = \boldsymbol{Z}^k_{2} \odot \exp(\boldsymbol{scale}) + \boldsymbol{tran}. \label{eqn: Z}
\end{equation}
The output of the first coupling layer is formed by concatenating the transformed half, $\tilde{\boldsymbol{Z}}^k_{2}$, with the unchanged half $\boldsymbol{Z}^k_{1}$.
The second coupling layer mirrors the first but with the roles of the two halves reversed. It takes the output of the first layer and applies the same which use $\tilde{\boldsymbol{Z}}^k_{2}$ to compute the scaling and translation factors and transforming $\boldsymbol{Z}^k_{1}$ accordingly, resulting in $\tilde{\boldsymbol{Z}}^k_{1}$.

Finally, the outputs from the two coupling layers are concatenated to produce the output of the $k$-th flow as Eqn. (\ref{eqn: FLow}) denotes.
\begin{equation}
\boldsymbol{Z}^{k+1} = Concat(\tilde{\boldsymbol{Z}}^{k}_1, \tilde{\boldsymbol{Z}}^{k}_2). \label{eqn: FLow}
\end{equation}
This process repeats for all $K_f$ flows, gradually transforming the Gaussian distribution into a complex non-parametric distribution.

By retaining the factors in coupling layers, the model enables efficient and bidirectional conversion between distributions, facilitating both inference and generation tasks. In summary, the carefully designed NF architecture, incorporating coupling layers and alternating transformations, enhances the model's expressiveness and flexibility. These advantages collectively contribute to the model's ability to effectively capture the complex latent space of cloud service indicators and improve overall performance.
\subsubsection{Update Parameters}
After obtaining the output of the last Normalizing Flow $\boldsymbol{Z}^{K_f +1} \in \mathbb{R}^{N}$ and the ground truth $\boldsymbol{X}_{T+1} \in \mathbb{R}^{N}$, the loss function is directly computed on the transformed samples. We use the mean squared error (MSE) loss in Eqn. (\ref{eqn: MSE}) to update the parameters of the Encoder and Decoder.
\begin{equation}
\mathcal{L} = \frac{1}{N} || \boldsymbol{X}_{T+1} - \boldsymbol{Z}^{K_f+1} ||_2^2. \label{eqn: MSE}
\end{equation}

The MSE loss function is chosen due to the absence of a parameterized probability distribution, making it impractical to apply loss functions based on probability density, such as the negative log-likelihood. It offers simplicity, statistical significance, contributing to faster optimization. In addition, it is important to note that alternative loss functions such as quantile loss could also be considered for their ability to handle extreme cases or outliers. After training process, we sample from the Gaussian distribution $\boldsymbol{D}_G$ and transform the samples through decoder to obtain  non-parametric distributions $\boldsymbol{D} \in \mathbb{R}^{N\times S}$ of indicators where $S$ is the number of samples. Leveraging the advantages of parameterized distributions and normalizing flows, the approach is valuable for producing samples that better match real-world distributions, thereby providing guidance for decision.  Alg. \ref{Alg: Training} summarizes the training process.

\section{Decision under Uncertainty}
\begin{figure}[htbp]
    \centering
    \includegraphics[width=\linewidth]{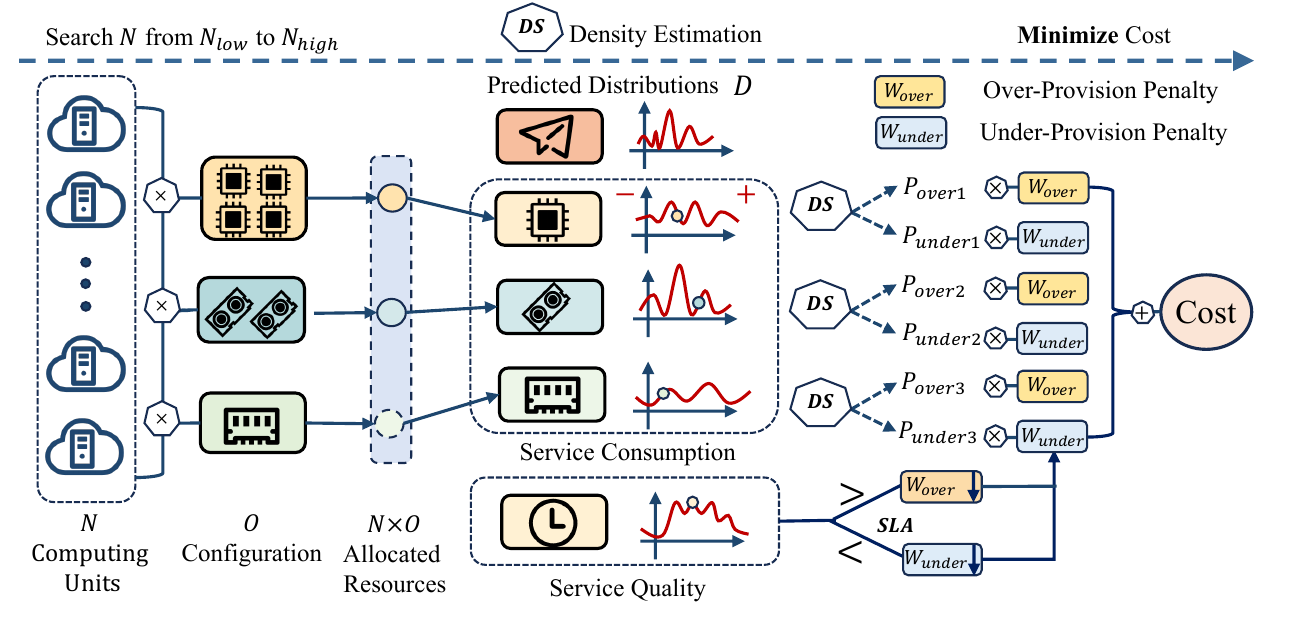}
    \caption{Bayesian Decision-Making Algorithm Design. The figure demonstrates the exploration of  $N \in [N_{low}, N_{high}]$ to find the optimal value that minimizes the total cost.}
    \label{fig:decision}
\end{figure}

As highlighted in \textbf{Challenge 3}, relying solely on point predictions for decision-making fails to account for the uncertainty in forecasts, which compromises decision stability and effectiveness. To address this issue, we designed a Bayesian decision-making method that utilizes the distribution obtained from our forecasting model. The Bayesian approach leverages the full predictive distribution, not just a single point estimate, allowing us to incorporate uncertainty into the decision-making process. For each potential decision, we compute the expected cost by combining the future distribution of the forecast indicators with the configuration of the computational units. Ultimately, we select the decision that minimizes the expected cost within the  range as the final decision. Fig. \ref{fig:decision} exhibits the framework of our decision process.

The input to our decision algorithm $\mathcal{A}$ comprises the forecasted distribution $\boldsymbol{D} \in \mathbb{R}^{N \times S}$ and the configuration of computational units $\boldsymbol{O} \in \mathbb{R}^{C}$. For simpler scenarios, the Gaussian distribution $\boldsymbol{D}_G$ produced by the encoder can be used instead. To ensure compatibility with the computational units, We use $\boldsymbol{D}_c \in \mathbb{R}^{C\times S}$ to represent the distributions of service consumption indicators.

Our objective is to determine the optimal number of computational units (CUs). We achieve this by searching within an operational range from $N_{low}$ to $N_{high}$. For each candidate number $N$, we calculate the corresponding cost and select the configuration with the minimum cost as our optimal decision.

For each $N \in [N_{low}, N_{high}]$, the cost is computed individually for each service consumption indicator. Specifically, we calculate the resources to be allocated as $N \times \boldsymbol{O}{c}$. We then assess the over-provisioning probability  $\boldsymbol{P}_{over}= \boldsymbol{P}(\boldsymbol{D}_c > N \times \boldsymbol{O}_c)$ and the under-provisioning probability $\boldsymbol{P}_{under} = \boldsymbol{P}(\boldsymbol{D}_c < N \times \boldsymbol{O}_c)$. These probabilities can be estimated by various methods such as kernel density estimation or by simply computing the proportion of samples in $\boldsymbol{D}_c$ that are greater or less than $N \times \boldsymbol{O}_c$. Understanding the likelihood of over-provisioning and under-provisioning, the decision can better manage risks related to resource allocation. It helps ensure that resources are neither wasted nor insufficient.

The cost for each indicator is derived by weighting the probabilities $P_{over}$ and $P_{under}$ with weights $W_{over} \in \mathbb{R}^{C}$ and $W_{under} \in \mathbb{R}^{C}$, respectively. The overall cost is then computed as Eqn. (\ref{eqn:cost}) denotes: 
\begin{equation}
Cost = W_{over} \cdot P_{over} + W_{under} \cdot P_{under}.
\label{eqn:cost}
\end{equation}

Finally, we select the number of CUs $N$ that minimizes the cost as the optimal number of CUs, denoted as $N_{optimal}$.

In continuous online decision-making, we dynamically adjust the weights $W_{over}$ and $W_{under}$ based on the estimated service quality indicators $D_p$. For example, if the mean of the predicted response time distribution approaches the Service Level Agreement (SLA) constraints, we reduce the weight of the over-provisioning penalty $W_{over}$. This adjustment biases the decision algorithm towards allocating more resources, thereby maintaining high service quality. Alg. \ref{Alg: Decision} details our Bayesian decision-making process.

In summary, our approach offers several key advantages:

\textbf{(1) Modeling Uncertainty}: By incorporating the full distribution of multiple indicators, our algorithm produces optimal decisions that avoid the limitations of point predictions.

\textbf{(2) Comprehensive Optimization}: Our algorithm can jointly optimize across different service indicators. It allows for a more balanced resource allocation in contrast to methods that rely on existing single-indicator decisions.

\textbf{(3) SLA Compliance and Adaptivity}: Our algorithm takes SLA constraints into account by dynamically adjusting the weights for over-provisioning and under-provisioning based on predicted service quality. It also adapts to a wide range of business scenarios by allowing users to prioritize the allocation of specific resources, such as GPUs, according to their needs through adjustable weights.
\begin{algorithm}[htbp]
\DontPrintSemicolon
\KwIn{Forecasting distribution $\boldsymbol{D}$, computational units configuration $\boldsymbol{O}$, operational range $[N_{low}\,, N_{high}]$, weights $W_{over}$ and  $W_{under}$.}
\KwOut{Optimal number of CUs $N_{optimal}$.}
Dynamically adjust the weights $W_{over}$ and $W_{under}$ based on service quality indicators $\boldsymbol{D}_p$;\;
$MinCost \leftarrow INF$;\;
\For{$N \in [N_{low}, N_{high}]$}{
    Calculate the resources to be allocated as $ N \times \boldsymbol{O}_c$;\;
    
    Compute the probability of over-provisioning $\boldsymbol{P}_{over} = \boldsymbol{P}(\boldsymbol{D}_c >N \times \boldsymbol{O}_c)$;\;
    Compute the probability of under-provisioning $\boldsymbol{P}_{under} = \boldsymbol{P}(\boldsymbol{D}_c < N \times \boldsymbol{O}_c)$;\;
    
    Calculate the cost for each indicator using Eqn.~\eqref{eqn:cost};\;
    \ForAll{$c \in C$}{
        $Cost_c = W_{over}[c] \cdot P_{over}[c] + W_{under}[c] \cdot P_{under}[c]$
    }
    \If{$Cost < MinCost$}{
        $MinCost \leftarrow Cost$;\;
        $N_{optimal} \leftarrow N$;
    }
}

\Return  $N_{optimal}$.\;

\caption{Bayesian Decision-Making}
\label{Alg: Decision}
\end{algorithm}

\section{Experiments}
In this section, we conduct a series of experiments designed to address the following research questions:

\textbf{RQ1:} How does the performance of \oursys compare with SOTA methods in predicting indicators of cloud service workload?

\textbf{RQ2:} Can the key components within \oursys be identified as significant contributors to its superior performance?

\textbf{RQ3:} In a real-world scenario of automatic scaling decision-making, can our approach achieve SOTA performance?

\subsection{Experimental Settings}

\subsubsection{Datasets}
Our study utilizes a comprehensive dataset from various types of cloud services to demonstrate the effectiveness of our scaling method. The dataset includes the following components:
(1) \textbf{Pay\_GPU}: This dataset consists of indicators collected from $100$ high-load GPU inference services over a one-week period, featuring $3441$ distinct GPUs. The data is sourced from the cloud platform that supports Alipay, one of the world's largest payment applications with over $1$ billion users globally.  
It represents a high-performance computing environment optimized for GPU-based tasks.
(2) \textbf{Pay\_CPU}: This dataset incorporates data from 100 CPU services with a total of $120,248$ CPU cores, also collected over a one-week period from Alipay. It represents a general-purpose cloud infrastructure supporting CPU-intensive operations.
(3) \textbf{Ali}\footnote{https://github.com/alibaba/clusterdata/tree/v2018}: Sourced from the Alibaba Cluster Trace, this dataset documents multivariate workload indicators from $50$ machines over an $8$-day period, showcasing a typical machine-based cloud architecture.
(4) \textbf{Fisher}\footnote{https://github.com/chrisliu1995/Fisher-model/tree/master}: This dataset includes workload data from $10$ containers over a $30$-day period within a Kubernetes framework, representing a modern containerized environment. These diverse datasets selection ensures \oursys's broad applicability and robust performance across different cloud service architectures.
\subsubsection{Data preprocessing}
In our data processing pipeline, we intentionally refrained from applying any preprocessing to sudden changes in the data, recognizing their prevalent occurrence within cloud services. For data normalization, we utilized MinMax normalization to standardize all indicators. Given the primary origin of our data from model inference services, we aggregated all indicators into $10$-minute intervals using the maximum value of each indicator for each interval to meet production environment requirements, enabling effective data management and the model's real-time prediction capabilities while addressing potential Service Level Agreement (SLA) violations resulting from under-provisioning.
Subsequently, the dataset was partitioned into training, testing, and validation sets in a 7:1:2 ratio to ensure robust model evaluation.
Following Def. \ref{Def: Indicators}, we categorize the service indicators into distinct groups such as \textbf{request}, \textbf{consumption}, and \textbf{quality}, generating Level Information as one of the inputs of \oursys.

\subsubsection{Baselines}
To evaluate the performance of \oursys, we compare it with $9$ state-of-the-art baselines. These baselines encompassed probabilistic prediction models: \textbf{DeepAR (2017)} \cite{deepar}, \textbf{CSDI (2021)} \cite{csdi}, \textbf{MG-TSD (2024)} \cite{mg-tsd}, and multivarate prediction methods: \textbf{Pyraformer (2022)} \cite{pyraformer}, \textbf{DLinear (2023)} \cite{dlinear},\textbf{PatchTST (2023)} 
 \cite{patchtst}, \textbf{Crossformer (2023)} 
 \cite{crossformer},\textbf{TSMixer (2023)} \cite{tsmixer}, \textbf{iTransformer (2024)} \cite{iTransformer}. Appx.~\ref{baseline} provides a detailed introduction.

\subsubsection{Evaluation Metrics}

To assess the accuracy of \oursys in cloud service prediction, we use two metrics: Mean Squared Error (MSE) and Continuous Ranked Probability Score (CRPS). MSE evaluates the accuracy of point predictions, measuring the average squared difference between predicted and actual values. CRPS, a more comprehensive metric, assesses the quality of probabilistic forecasts. It evaluates how well the entire predicted probability distribution aligns with the observed outcome, considering both accuracy and forecast confidence. Lower CRPS scores indicate better probabilistic predictions. For probabilistic methods, we sample from the predicted distribution to compute MSE. For point prediction methods, we repeat the point predictions as inputs for CRPS calculation. By using both MSE and CRPS, we provide a thorough assessment of \oursys's performance in both point and probabilistic predictions for cloud service contexts.
\subsubsection{Parameter Settings}

\oursys is developed using PyTorch and trained on an NVIDIA P100 GPU. The hyper parameter settings are as follows: the learning rate is set to 0.001, the batch size is set to $128$, and the dimension of the representation $d$ is set to $512$. The numbers of Hierarchical Attention block and Normalizing Flow are both set to $2$. All models are trained for 20 epochs, and the best-performing model on the validation set is utilized for testing.

\begin{table*}[htbp]
\label{Table_main}

\caption{Comparisons Results on four datasets. The best results are in boldface and the second best results are underlined. }

\scalebox{0.9}{
\begin{tabular}{ccccccccccccc}
\toprule
\multicolumn{1}{l}{Datasets} & T & Metrix & DeepAR & CSDI & MG-TSD & Pyraformer & DLinear & PatchTST & Crossformer & TSMixer & iTransformer & \oursys  \\ \toprule
\multirow{6}{*}{Pay\_GPU} & \multirow{2}{*}{72} & MSE & 0.0225 & 0.0374 & 0.0331 & 0.0227 & 0.0350 & 0.0202 & 0.8261 & 0.2013 & \underline{0.0179} & \textbf{0.0171}  \\
 &  & CRPS & \underline{0.0706} & 0.1214 & 0.1454 & 0.0781 & 0.1176 & 0.0719 & 0.6652 & 0.3686 & 0.0708 & \textbf{0.0563}  \\ \cline{2-13} 
 & \multirow{2}{*}{144} & MSE & 0.0291 & 0.0409 & 0.0357 & \underline{0.0241} & 0.0419 & 0.0267 & 1.2853 & 0.0880 & 0.0248 & \textbf{0.0229}   \\
 &  & CRPS & \underline{0.0807} & 0.1302 & 0.2628 & 0.0859 & 0.1372 & 0.0893 & 0.9662 & 0.2160 & 0.0892 & \textbf{0.0664}   \\ \cline{2-13} 
 & \multirow{2}{*}{288} & MSE & \underline{0.0366} & 0.0513 & 0.0443 & 0.0374 & 0.0413 & 0.0481 & 1.6736 & 0.5510 & 0.0476 & \textbf{0.0355}   \\
 &  & CRPS & \underline{0.0992} & 0.1828 & 0.3062 & 0.1140 & 0.1323 & 0.1363 & 1.1406 & 0.5487 & 0.1328 & \textbf{0.0890}  \\ \toprule
\multirow{6}{*}{Pay\_CPU} & \multirow{2}{*}{72} & MSE & 0.3593 & 0.2674 & 0.1258 & 0.0676 & 0.1365 & \underline{0.0327} & 0.3639 & 0.2997 & 0.0337 & \textbf{0.0323}  \\
 &  & CRPS & 0.2142 & 0.5457 & 0.3317 & 0.0891 & 0.1920 & \underline{0.0708} & 0.3067 & 0.2347 & 0.0711 & \textbf{0.0572} \\ \cline{2-13} 
 & \multirow{2}{*}{144} & MSE & 0.4260 & 0.2680 & 0.1718 & 0.1084 & 0.1748 & \underline{0.0604} & 0.4071 & 0.2465 & 0.0683 & \textbf{0.0448}  \\
 &  & CRPS & 0.2170 & 0.5273 & 0.4722 & 0.1146 & 0.2066 & \underline{0.0921} & 0.3866 & 0.2449 & 0.0962 & \textbf{0.0730}   \\ \cline{2-13} 
 & \multirow{2}{*}{288} & MSE & 0.4997 & 0.2998 & 0.2196 & 0.1441 & 0.2442 & 0.1219 & 0.5541 & 0.1610 & \underline{0.1103} & \textbf{0.0762}   \\
 &  & CRPS & 0.2351 & 0.5369 & 0.5837 & 0.1519 & 0.2356 & 0.1298 & 0.4622 & 0.2060 & 0.1295 & \textbf{0.0943}  \\ \toprule
\multirow{6}{*}{Ali} & \multirow{2}{*}{72} & MSE & 0.0044 & 0.0172 & 0.0318 & \underline{0.0042} & 0.0112 & 0.0044 & 0.0891 & 0.0410 & 0.0047 & \textbf{0.0039}   \\
 &  & CRPS & 0.0399 & 0.0730 & 0.0535 & \underline{0.0327} & 0.0821 & 0.0328 & 0.2359 & 0.1550 & 0.0340 & \textbf{0.0289}   \\ \cline{2-13} 
 & \multirow{2}{*}{144} & MSE & 0.0054 & 0.0116 & 0.0323 & 0.0056 & 0.0145 & \underline{0.0049} & 0.0962 & 0.0246 & 0.0055 & \textbf{0.0048}  \\
 &  & CRPS & 0.0492 & 0.0588 & 0.0729 & 0.0401 & 0.0855 & 0.0351 & 0.2496 & 0.1235 & 0.0390 & \textbf{0.0290}  \\ \cline{2-13} 
 & \multirow{2}{*}{288} & MSE & 0.0066 & 0.0134 & 0.0349 & 0.0062 & 0.0167 & \textbf{0.0058} & 0.0612 & 0.0239 & 0.0071 & \underline{0.0061}  \\
 &  & CRPS & 0.0542 & 0.0613 & 0.0852 & 0.0473 & 0.0888 & \underline{0.0426} & 0.2048 & 0.1166 & 0.0482 & \textbf{0.0399} \\ \toprule
\multirow{6}{*}{Fisher} & \multirow{2}{*}{72} & MSE & 0.0695 & 0.0766 & 0.0650 & \underline{0.0461} & 0.0500 & 0.0467 & 0.0587 & 0.0581 & 0.0470 & \textbf{0.0453}   \\
 &  & CRPS & 0.1481 & 0.1634 & 0.1840 & \underline{0.1141} & 0.1448 & 0.1143 & 0.1545 & 0.1484 & 0.1160 & \textbf{0.0995}  \\ \cline{2-13} 
 & \multirow{2}{*}{144} & MSE & 0.0604 & 0.0968 & 0.0663 & \underline{0.0469} & 0.0574 & 0.0472 & 0.0616 & 0.0863 & 0.0489 & \textbf{0.0458}   \\
 &  & CRPS & 0.1390 & 0.2282 & 0.1991 & 0.1172 & 0.1606 & \underline{0.1159} & 0.1596 & 0.1691 & 0.1200 & \textbf{0.0940}\\ \cline{2-13} 
 & \multirow{2}{*}{288} & MSE & 0.0562 & 0.0934 & 0.0757 & \underline{0.0483} & 0.0557 & 0.0485 & 0.0640 & 0.1046 & 0.0496 & \textbf{0.0474}  \\
 &  & CRPS & 0.1276 & 0.1481 & 0.2057 & 0.1213 & 0.1450 & \underline{0.1195} & 0.1594 & 0.1903 & 0.1237 & \textbf{0.1097}   \\\toprule
 
\multicolumn{1}{l}{\multirow{2}{*}{Avg. Rank}} & \multirow{2}{*}{All} & MSE &  5.9167 & 7.8333 & 6.7500  & 3.1667 & 6.0833 & \underline{3.0833}  & 9.1667 &  8.0833 & 3.7500  & \textbf{1.0000}    \\
\multicolumn{1}{l}{} &  & CRPS & 4.5833  & 7.8333 & 8.5000  & 3.3333 & 6.3333  & \underline{3.0000} & 8.8333  & 8.0000 & 3.5833  &  \textbf{1.0833}  \\ \bottomrule
\end{tabular}
}
\label{Table_main}
\end{table*}

\subsection{\textbf{RQ1}: Comparative Results} \label{sec: comp}

Table \ref{Table_main} presents the comparative results between \oursys and various baselines. The experimental data lead to several insights:

(1) \oursys demonstrates superior performance in both point prediction and probabilistic prediction across four diverse cloud service datasets, achieving optimal results with average improvements of $7.84$\% in MSE and $15.9$\% in CRPS. These consistent improvements across varied datasets underscore \oursys's generalization capability, evidencing its efficacy in providing accurate predictions for cloud services with different architectures. We attribute the enhancement in point prediction primarily to the hierarchical attention mechanism, which effectively models the causal relationships and sequential dependencies among indicators, thus addressing \textbf{Challenge 1}. The improvement in distribution prediction can be attributed to the Normalizing Flow's non-parametric modeling of cloud service indicator distributions, addressing \textbf{Challenge 2}. 

(2) The results from the majority of models reveal a counter intuitive trend: for cloud services characterized by short adjustment cycles and rapidly evolving indicators, longer input length $T$ do not consistently lead to improved prediction accuracy. This phenomenon can be elucidated by considering the dynamic nature of cloud services. Longer input sequences inevitably incorporate a greater number of historical time patterns, some of which may have become obsolete due to recent service adjustments or evolving operational conditions. These outdated patterns, when included in the prediction process, can introduce noise and potentially skew the model's understanding of current trends, thereby compromising prediction accuracy.
Consequently, when developing predictive models for cloud service indicators, it is imperative to calibrate input sequence length based on service-specific characteristics.

(3) Attention-based methods exhibit stronger generalization capabilities, providing adaptive performance across datasets. With the exception of Crossformer, whose performance may be sensitive to segment size configurations due to its locality-based embedding approach, attention-based methods, including Pyraformer, PatchTST, and iTransformer, consistently demonstrate superior performance across all datasets, ranking second only to \oursys. While MLP-based approaches (TSMixer, DLinear) show limitations in fitting complex, large-scale data, as evidenced by their poor performance on \textbf{Pay\_GPU} datasets. Similarly, CSDI's inability to converge on the \textbf{Pay\_CPU} dataset suggests that the multiple iterations of Gaussian noise addition and denoising in diffusion process may exacerbate the challenge of fitting the indicators. 

\begin{figure*}[htbp]
\centering
\includegraphics[width=0.98\linewidth]{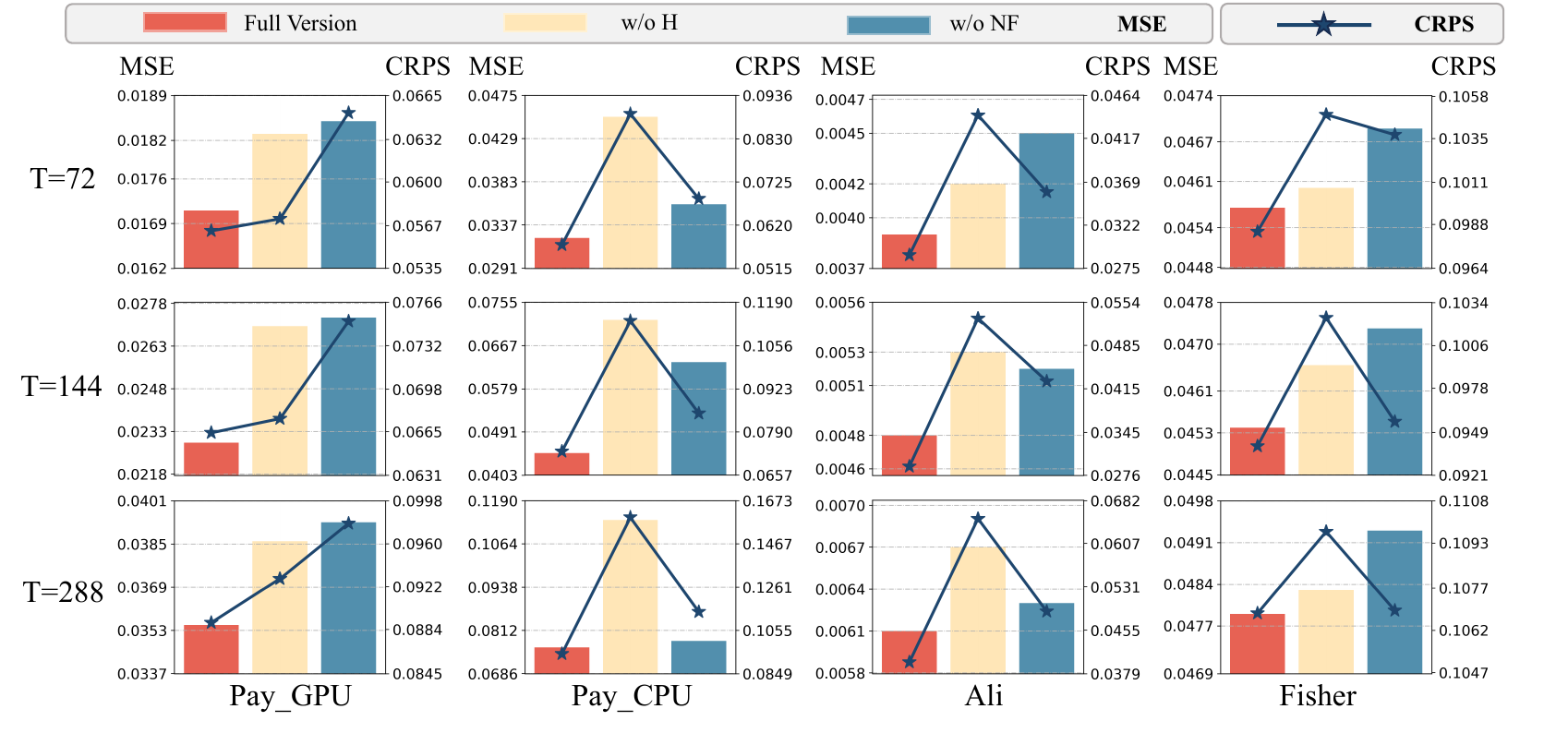}
\caption{Ablation Study. Comparison of Prediction Accuracy between \oursys and its variant on MSE and CRPS respectively.}
\label{fig:ABStudy}
\end{figure*} 
\subsection{RQ2: Ablation Study}
We conducted ablation studies to analyze the impact of key components within our system. The outcomes are shown in Fig. ~\ref{fig:ABStudy}.

(1) Impact of Hierarchical Attention: We replace the Hierarchical Attention with Full Attention blocks that do not differentiate levels and computing the attention coefficients for all indicators pairwise. This substitution is denoted as 'w/o H'. Experimental results indicate that such a replacement leads to a deterioration in both MSE and CRPS across all input lengths in all cloud service datasets. Notably, for the \textbf{CPU} service, which exhibits more consistent and stable relationships among indicators, the removal of Hierarchical Attention results in an average increase of $49.7\%$ in Mean Squared Error (MSE). We attribute it to the Full Attention Mechanism's failure to account for the hierarchical relationships between indicators, leading to confusion in information propagation across different layers. Thus, for cloud service with clear structures, Hierarchical Attention proves to be a more effective choice.

(2) Impact of Normalizing Flow: We compared our model with a variant using a simple linear layer for Gaussian distribution mapping, denoted as 'w/o NF'. Results show significant increases in both Mean Squared Error (MSE) and Continuous Ranked Probability Score (CRPS) for the 'w/o NF' variant. This suggests that the Normalizing Flow better captures the true data distribution, effectively addressing \textbf{Challenge 2}. The improvement is particularly pronounced in complex cloud environments like \textbf{Pay\_GPU}, \textbf{Ali}, and \textbf{Fisher}, where operational patterns are intricate and subject to rapid changes. In these scenarios, the Normalizing Flow demonstrates superior adaptability and fidelity to the underlying distribution. Thus, incorporating the Normalizing Flow substantially enhances performance in dynamic service environments, underscoring its importance in accurate cloud service prediction.

\subsection{RQ3: Online Decision-Making }
We conducted a comprehensive one-month online A/B test from June 12, 2024, on Alipay's internal cloud service inference platform. The test focused on ten large-scale GPU services, evaluating the performance of \oursys in predicting future workload indicators and making resource allocation decisions using Algorithm \ref{Alg: Decision}.  Appx. \ref{sec: deploy} provides detailed information on the deployment process.

The key metrics we considered included CPU and GPU usage hours ($CPU\_Usage$, $GPU\_Usage$), resource utilization $(GPU\_Uti$, $GPU\_Uti)$, and the probability of allocated resources exceeding actual consumption $SuccR$. Higher resource utilization rates while maintaining a high $SuccR$ signifies better performance of the scaling method. We compared our approach with the following methods:
The first group is Non-Predictive Methods:
\textbf{Rule-Based}: A method uses the max of historical indicators to make resource allocations.
\textbf{Autopilot (2020)} \cite{autopilot}: A classical auto-scaling method that selects the maximum value from historical indicators over a given period and updates periodically.
To avoid SLA violations, a 10\% buffer is added to the Non-Predictive Methods.
Predictive Methods: \textbf{FIRM (2020)}  \cite{firm}: An adaptive auto-scaling method using reinforcement learning to make decisions based on service quality indicators. \textbf{FSA (2023)} \cite{IJCAI-1}: A method uses service request predictions to make resource allocation decisions, taking into account the relationship between service request and service consumption.

\begin{figure}[htbp]
    \centering
    \subfigure[CPU]{
        \centering
        \includegraphics[width=0.45\linewidth]{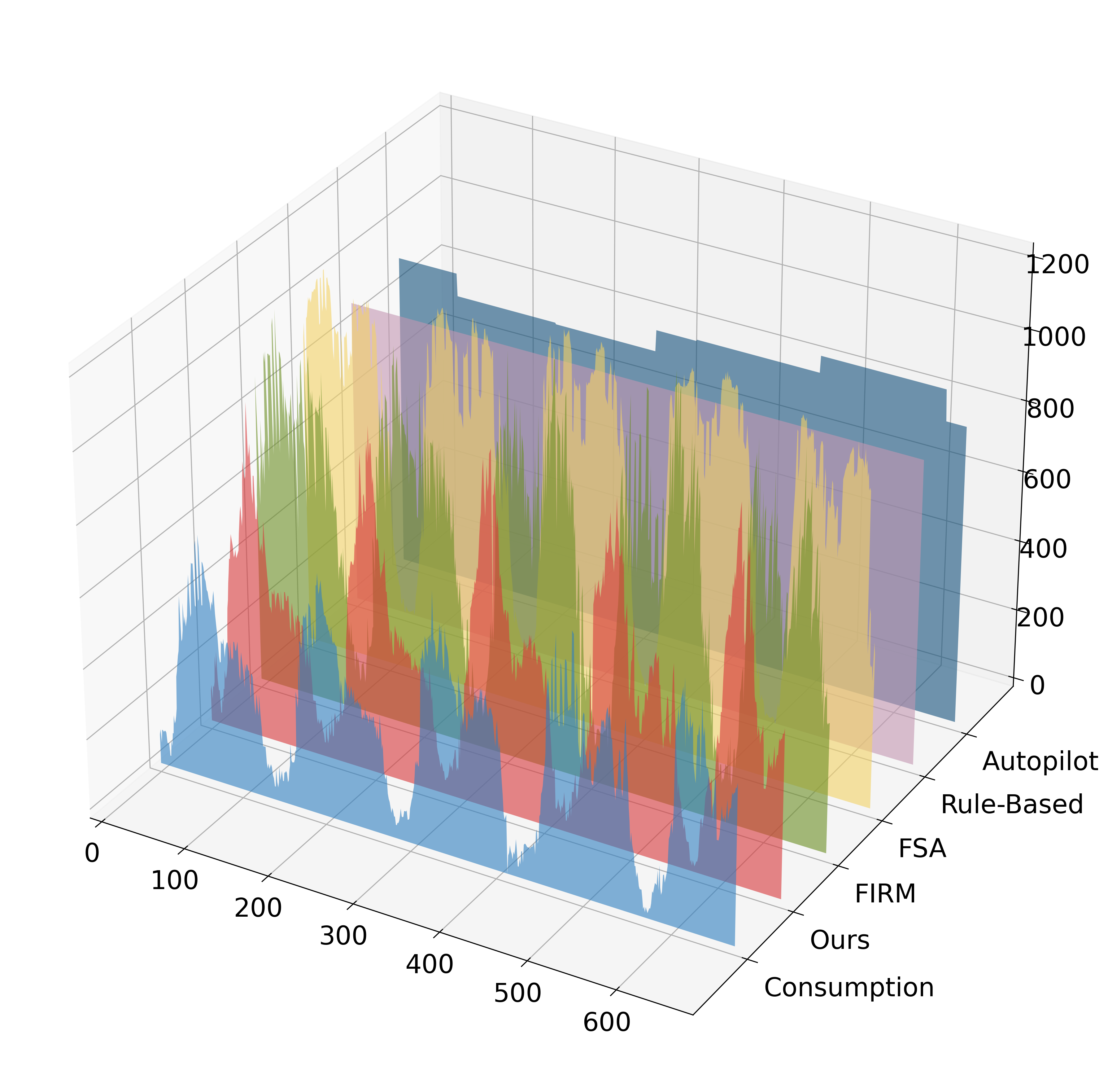}
        \label{fig:CPU}
        }
    \subfigure[GPU]{
        \centering
        \includegraphics[width=0.45\linewidth]{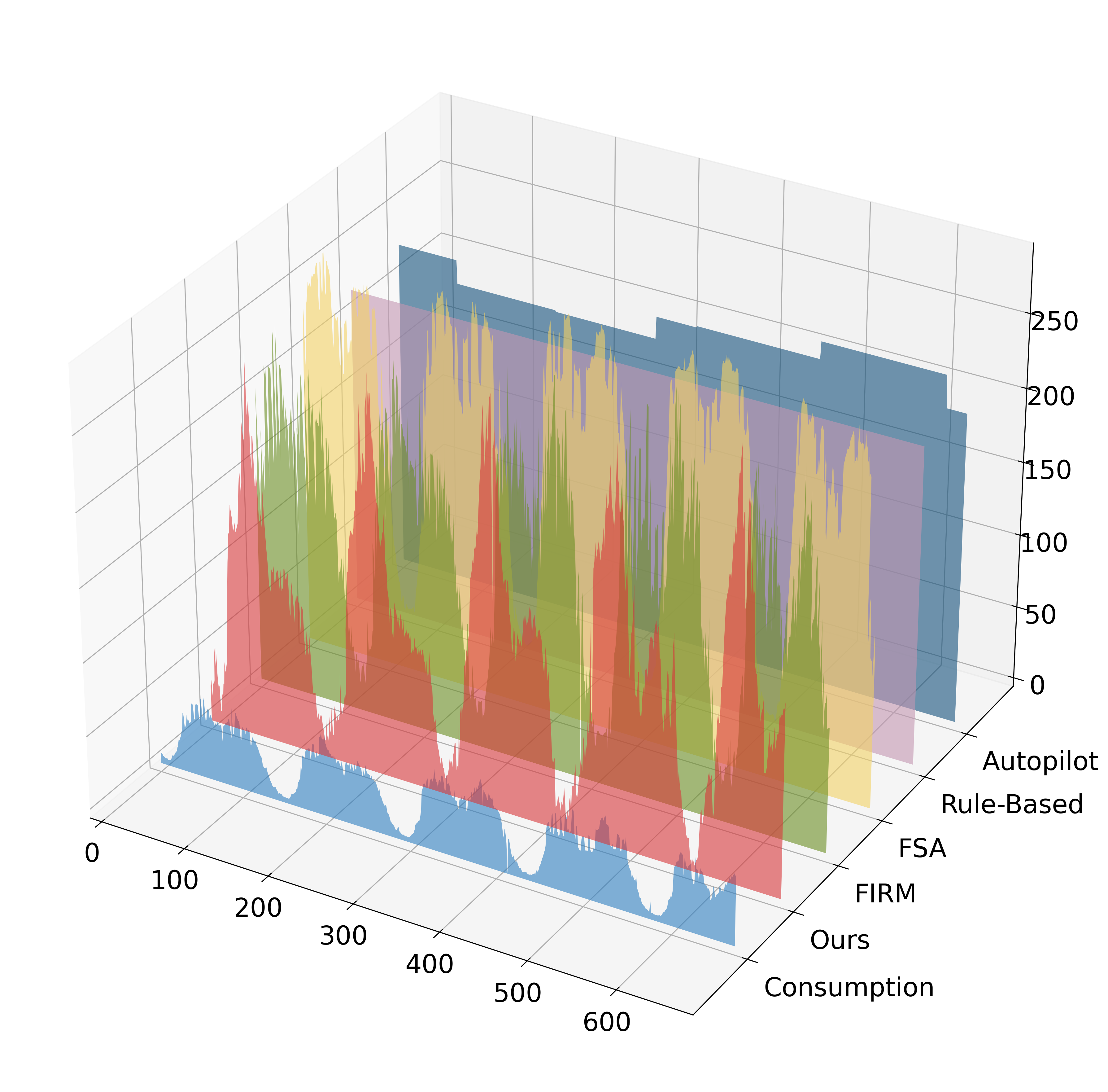}

        \label{fig:GPU}
    }
    \caption{Comparison of Decision Results. For a same GPU service, left and right subfigures show allocated CPU and GPU resources respectively from different methods over time stamps (X-axis). Z-axis represents allocated resources. "Consumption" denotes actual service resource usage.}
    \label{fig:Decision}
\end{figure}

From Fig. \ref{fig:Decision}, it is evident that our approach adeptly aligns resource allocation with demand fluctuations, thereby allocating fewer resources compared to other methods during peak and trough periods. Table \ref{Table:Decision} further substantiates our method's superior performance, achieving significantly higher utilization rates while maintaining a high success rate (SuccR) of 99.82\%. This underscores the efficacy of our Bayesian decision-making algorithm, which incorporates multiple indicators to model uncertainties, thus achieving an optimal balance between resource savings and SLA constraints. The marginally lower SuccR can be attributed to occasional service restarts, which could be addressed through proactive human intervention. Notably, our approach conserves $240,000$ CPU core hours and $35,000$ GPU hours over the month-long test period compared to the second-best method, translating to estimated economic benefits exceeding $\$100,000$. This substantial cost saving is achieved while ensuring high service quality. Overall, our evaluation demonstrates the effectiveness of \oursys in optimizing resource allocation, enhancing resource utilization, achieving significant cost savings, and maintaining high service quality.

\begin{table}[htbp]
\caption{Online Decision Results}
\label{Table:Decision}
\scalebox{0.8}{
\begin{tabular}{cccccc}
\toprule
Methods & CPU\_Usage & GPU\_Usage & CPU\_Uti & GPU\_Uti & SuccR  \\
\toprule
NoScale & 1.02e+7      & 1.13e+6   & 16.27\%  & 18.66\% & 100\%\\
Rule-Based & 5.56e+6  & 6.50e+5  & 29.87\%   & 32.46\% & 99.92\%  \\
Autopilot & 5.35e+6 & 6.17e+5  & 31.03\%  & 34.20\% & 99.88\% \\
FIRM & 3.72e+6  & 4.09e+5  & 44.62\%   & 51.59\% & 99.23\%   \\
FSA & 3.10e+6    & 3.54e+5 & 53.55\%   &    59.60\% & 95.87\%             \\
\hline
\oursys & \textbf{2.86e+6}  & \textbf{3.19e+5 } & \textbf{58.04\%} & \textbf{66.14\%} & 99.82\%\\ 
\bottomrule

\end{tabular}}
\end{table}

\section{Conclusion}
We introduce \oursys, a novel framework addressing cloud service resource scaling challenges by integrating multi-indicator distribution predictions into a Bayesian decision process. \oursys leverages a hierarchical attention mechanism and Normalizing Flows to capture hierarchical dependencies and non-Gaussian distributions, ensuring accurate and reliable resource predictions. Empirical validation showed that \oursys outperforms established methods and delivers significant economic benefits, saving over $35,000$ GPU computation hours in a month-long A/B test, demonstrating its practical impact in large-scale cloud service environments.

\section{ACKNOWLEDGEMENT}
This work was sponsored by the National Natural Science Foundation of China [U23A20309, 62272302, 62172276, 62372296], CCF-Ant Research Fund [CCF-AFSG RF20230408] and the Shanghai Municipal Science Technology 
Major Project [2021SHZDZX0102].
\bibliographystyle{ACM-Reference-Format}
\bibliography{sample-base}

\appendix

\section{Notation Table \& Pesudocode}

\begin{table}[H]
\caption{Summary of Primary Symbols}
\label{tab:symbol}
\scalebox{0.9}{
\begin{tabular}{c|c}
\toprule
\textbf{Symbol} & \textbf{Description} \\
\toprule
$\boldsymbol{X} \in \mathbb{R}^{N \times T} $ & Historical indicators \\
$\boldsymbol{L} \in \mathbb{Z}^{N}$ & Level information of indicators\\
$\boldsymbol{O} \in \mathbb{Z}^{C}$ & Computational unit's configuration  \\
$\boldsymbol{E} \in \mathbb{R}^{(N + F) \times D}$ & Enhanced Indicator Embedding.\\
$\boldsymbol{H}^{k+1} \in \mathbb{R}^{(N + F) \times D}$ &  Embeddings after Hierachical Attention\\
$\boldsymbol{D}_G \in \mathbb{R}^{N\times 2} $ & Learned Latent Gaussian Distributions \\
$\boldsymbol{Z}^{k+1} \in \mathbb{R}^{N} $  & Transformed samples of indicators\\
$\boldsymbol{D} \in \mathbb{R}^{N \times T} $ & Predicted  future  distribution of indicators \\
$N_{optimal}$ & Optimal number of CUs  after decision algorithm\\
\bottomrule
\end{tabular}}
\end{table}

\begin{algorithm} [h] 
\DontPrintSemicolon
\KwIn{Historical Indicators $\boldsymbol{X}$, Level Information $\boldsymbol{L}$, number of training epochs $J$, learning rate $r$.}
\KwOut{Model Parameters $\boldsymbol{\Theta}$, Learned Distributions.}
\For{$j = 1 \textbf{ to } J$}{
    Generate the embedding of input indicators $\boldsymbol{E}$;\;

    Capture indicators' dependencies with hierarchical attention $\boldsymbol{H}^{k+1}$ based on the embedding $\boldsymbol{E}$ and level information $\boldsymbol{L}$ according to Eqn.~\eqref{eqn: M} - Eqn.~\eqref{eqn: FFN};\;

    Project the learned embedding $\boldsymbol{H}^{k+1}$ into a Gaussian Distribution $\boldsymbol{D}_G$ by Eqn.~\eqref{eqn: E};\;

    Sample from the Gaussian Distribution $\boldsymbol{D}_G$ and transform them by Normalizing Flows to get the transformed samples $\boldsymbol{Z}^{K_f+1}$ by Eqn.~\eqref{eqn: MLP} - Eqn.~\eqref{eqn: FLow};\;

    \For{$\theta \in \boldsymbol{\Theta}$}{
      $\theta \leftarrow \theta - r \cdot \nabla_\theta \mathcal{L}$;
    }
}

Sample from the trained $\boldsymbol{D}_G$ and transform them by NFs to get the non-parametric distribution $\boldsymbol{D}$;\;

\textbf{Return}  $\boldsymbol{\Theta}$, $\boldsymbol{D}_G$,  $\boldsymbol{D}$;\;

\caption{Training Process}
\label{Alg: Training}
\end{algorithm}

\section{Related Works} \label{sec: RW}
Existing research on cloud service resource scaling focuses on two aspects: \textbf{Workload Forecasting} and \textbf{Decision-Making}.

\textbf{Workload Forecasting:} Initial studies primarily targeted the prediction of single service request indicators \cite{IPOC, kaeinformer, IJCAI-1,Yang}. For instance, IPOC \cite{IPOC} designed an online ensemble algorithm to handle the variable temporal patterns of cloud service requests. KAE-Informer \cite{kaeinformer} combined time series decomposition and attention mechanisms for diverse load prediction tasks. These methods focused on the irregularities within individual indicators but did not account for the intrinsic correlations among various indicators. With the advancement of multivariate time series prediction methods, more approaches \cite{KDD24, group, iTransformer} have started to consider the relationships among multiple cloud service indicators. For example, STAMP \cite{KDD24} employed spatio-temporal graph neural networks, and GROUP \cite{group} used rule-based methods to capture inter-indicator correlations. However, these methods overlook hierarchical structures and causal relationships, thus failing to address \textbf{Challenge 1}. 

In recent years, the increased scale and complexity of cloud services have led to some works \cite{WSDM, Magic, uncertainty, csdi, mg-tsd} focusing on probabilistic distribution predictions instead of point predictions to tackle the uncertainty in indicator predictions described in \textbf{Challenge 3}. For example, MagicScaler \cite{Magic} predicts Gaussian distributions of workloads. However, these methods do not address complex distributions of service indicators, failing to meet \textbf{Challenge 2}.

\textbf{Decision in Resource Scaling}: In terms of scheduling, many methods use simple rule-based approaches or optimization modeling to convert prediction outcomes into Computational Unit decisions \cite{IPOC, KDD24, Magic, IJCAI-1, Robust}. For instance, IPOC \cite{IPOC} multiplies predicted service consumption by a coefficient for resource decisions, and RobustScaler \cite{Robust} integrates service quality and service consumption for optimization-oriented decision. However, these methods do not effectively utilize prediction uncertainty or achieve flexible, cross-indicator optimization, thus failing to address \textbf{Challenge 3}.

In summary, \oursys aims to fill the existing research gap in large-scale cloud services by integrating predictions of the distributions of multiple indicators and embedding them into the Bayesian decision-making process for joint optimization.

\section{Baseline Description} \label{baseline}
The first group consists of probabilistic prediction models. Among them, DeepAR is implemented by a standard library\footnote{https://ts.gluon.ai/stable/api/gluonts/gluonts.mx.model.deepar.html} while both CSDI and MG-TSD employed their official implementations. 

\textbf{DeepAR}: A commonly adopted probabilistic forecasting approach involves utilizing recurrent neural networks to project predictions onto a Gaussian distribution.

\textbf{CSDI}: A method utilzes score-based diffusion model to generate conditional probability predictions.

\textbf{MG-TSD}: An advanced diffusion-based model that leverages the inherent granularity levels within the data as targets at  diffusion steps, effectively guiding the learning process of diffusion models.

The second group is multivariate forecasting methods, which are the most widely used cloud service workload prediction methods. All the following baselines are implemented by open-source code\footnote{ https://github.com/thuml/Time-Series-Library}.

\textbf{Pyraformer}: A method utilizes pyramidal attention to effectively capture both short and long-term temporal dependencies with low time and space complexity.

\textbf{DLinear}: A method utilizing linear layers to capture temporal correlation for multivariate time-series prediction tasks.

\textbf{PatchTST}: A Transformer-based model designed for multivariate time series forecasting and self-supervised representation learning, utilizing a channel-wise independent mechanism.

\textbf{Crossformer}: A Transformer-based model that employs Dimension Segment Wise (DSW) embedding to capture cross-time and cross-dimension dependencies for multivariate series forecasting.

\textbf{TSMixer}: A novel architecture based on MLPs that efficiently captures the complexity of  multivariate series.

\textbf{iTransformer}:  A Transformer-based model that redefines the attention and feed-forward network to function on inverted dimensions, enhancing its ability to capture multivariate correlations.

\section{Deployment Details}
\label{sec: deploy}
To ensure reliability and minimize the risk of SLA violations, we have implemented several key practices for all prediction-based methods during deployment:

\textbf{90-90 Principle}: A critical step in our approach is the rigorous use of the 90-90 validation check before deploying any service. This practice ensures that at least $90\%$ of the predictions demonstrate an accuracy exceeding $90\%$. It is important to note that all $10$ services included in our testing have passed the 90-90 validation check. This practice significantly mitigates the risk of SLA breaches. Importantly, we find larger-scale inference services with more stable workload patterns have a higher likelihood of passing the 90-90 test. In practice, we focus on optimizing a subset of larger-scale services to achieve rapid benefits, rather than optimizing all services.

\textbf{Frequent Model Retraining}: Our model undergoes daily retraining in the production environment, and updates are deployed only after meeting the 90-90 criterion in offline testing to ensure consistent performance and adaptability to changing workloads. Our emphasis lies in prioritizing prediction accuracy over the training cycle, given the frequent adjustments in cloud service workloads. By opting for daily model retraining, we aim to avoid the influence of past, dissimilar workload patterns and improve prediction accuracy, as validated in Sec. \ref{sec: comp}. Moreover, the one-day model update cycle for a single cloud service provides sufficient time for retraining and enhances our focus on prediction accuracy improvement.

\textbf{Resource Allocation Indicators}: For a more direct comparison, we have primarily focused on optimizing CPU and GPU utilization as well as response time indicators while adjusting the weights in Bayesian decision-making.  Additionally, \oursys can accommodate joint optimization of memory, GPU memory, and other indicators as needed, adapting to utilize different combinations of indicators based on the specific demands of the task. 

These practices have been invaluable in overcoming data collection, modeling, and deployment challenges within a production environment. They provide a robust framework ensuring successful implementation of our resource scaling solution.
\end{document}